\documentstyle[eaclap]{article}

\newcommand{\myvalue}[1]{{\mbox{\normalsize{#1}}}}
\newcommand{\avm}[1]{{\setlength{\arraycolsep}{0.8mm}
                       \renewcommand{\arraystretch}{1.3}
                       \left[
                       \begin{array}{l}
                       \\[-3mm] #1 \\[-3mm] \\
                       \end{array}
                       \right]
                    }}
\newcommand{\atvl}[2]{{\mbox{\small{\tt #1}}{#2}}}
\newcommand{\alexcata}[3]{{\scriptsize \mbox{$
                            {\avm{\\[-3mm] {#1} \\ \\[-3mm]
                                           \atvl{l}{#2} \\ \\[-3mm]
                                           \atvl{r}{#3} \\[-3mm]}}
                                            $}}}
\newcommand{\alexcath}[4]{{\scriptsize \mbox{$
                            {\avm{\\[-3mm] {#1} \\ \\[-3mm]
                                           \atvl{l}{#2} \\ \\[-3mm]
                                           \atvl{r}{#3} \\ \\[-3mm]
                                           \atvl{h}{#4} \\[-3mm]}}
                                            $}}}
\newcommand{\alexcatw}[4]{{\scriptsize \mbox{$
                            {\avm{\\[-3mm] {#1} \\ \\[-3mm]
                                           \atvl{l}{#2} \\ \\[-3mm]
                                           \atvl{r}{#3} \\ \\[-3mm]
                                           \atvl{w}{#4} \\[-3mm]}}
                                            $}}}

\title{\vspace*{-17mm}
{\bf {\footnotesize In Proceedings of the 7th Conference of the
    European
Chapter of the ACL, EACL 95, Dublin, Ireland.} \\[3mm]
Incremental Interpretation of Categorial
Grammar\thanks{This research was supported by the U.K.\
Science and Engineering
Research Council, grant RR30718. I am grateful to Patrick Sturt, Carl Vogel,
and the reviewers for comments on an earlier version.}}}
\author{David Milward\\
Centre for Cognitive Science\\
University of Edinburgh\\
2 Buccleuch Place, Edinburgh, EH8 9LW, U.K. \\
davidm@cogsci.ed.ac.uk}

\begin{document}

\maketitle
\vspace{-0.5in}
\begin{abstract}

  The paper describes a parser for Categorial Grammar which provides
  fully word by word incremental interpretation. The parser does not
  require fragments of sentences to form constituents, and thereby
  avoids problems of spurious ambiguity. The paper includes a brief
  discussion of the relationship between basic Categorial Grammar and
  other formalisms such as HPSG, Dependency Grammar and the Lambek
  Calculus. It also includes a discussion of some of the issues which
  arise when parsing lexicalised grammars, and the possibilities for
  using statistical techniques for tuning to particular languages.

\end{abstract}

\section{Introduction}

There is a large body of psycholinguistic evidence which suggests that
meaning can be extracted before the end of a sentence, and before the
end of phrasal constituents (e.g.\ Marslen-Wilson 1973, Tanenhaus et
al.\ 1990). There is also recent evidence suggesting that, during
speech processing, partial interpretations can be built extremely
rapidly, even before words are completed (Spivey-Knowlton et al.\
1994)\footnote{Spivey-Knowlton et al.\ reported 3 experiments. One
  showed effects before the end of a word when there was no other
  appropriate word with the same initial phonology. Another showed
  on-line effects from adjectives and determiners during noun phrase
  processing.}.  There are also potential computational applications
for incremental interpretation, including early parse filtering using
statistics based on logical form plausibility, and interpretation of
fragments of dialogues (a survey is provided by Milward and
Cooper, 1994, henceforth referred to as M\&C).

In the current computational and psycholinguistic literature there are
two main approaches to the incremental construction of logical forms.
One approach is to use a grammar with `non-standard'
constituency, so that an initial fragment of a sentence, such as {\it
  John likes}, can be treated as a constituent, and hence be assigned
a type and a semantics.  This approach is exemplified by Combinatory
Categorial Grammar, CCG (Steedman 1991), which takes a basic CG with
just application, and adds various new ways of combining elements
together\footnote{Note that CCG doesn't provide a type for all initial
  fragments of sentences. For example, it gives a type to {\it John
    thinks Mary}, but not to {\it John thinks each}. In contrast the
  Lambek Calculus (Lambek 1958) provides an infinite number of types
  for any initial sentence fragment.}.  Incremental interpretation can
then be achieved using a standard bottom-up shift reduce parser,
working from left to right along the sentence.  The alternative
approach, exemplified by the work of Stabler on top-down parsing
(Stabler 1991), and Pulman on left-corner parsing (Pulman 1986) is to
associate a semantics directly with the partial
structures formed during a top-down or left-corner parse. For example,
a syntax tree missing a noun phrase, such as the following
\begin{verbatim}
             s
            / \
          np  vp
        John  / \
             v   np^
           likes
\end{verbatim}
can be given a semantics as a function from entities to truth values
i.e. {\bf $\lambda$x. likes(john,x)}, without having to say that {\it
John likes} is a constituent.

Neither approach is without problems. If a grammar is augmented with
operations which are powerful enough to make most initial fragments
constituents, then there may be unwanted interactions with the rest of
the grammar (examples of this in the case of CCG and the Lambek
Calculus are given in Section 2). The addition of extra operations
also means that, for any given reading of a sentence there will
generally be many different possible derivations (so-called `spurious'
ambiguity), making simple parsing strategies such as shift-reduce
highly inefficient.

The limitations of the parsing approaches become evident when we consider
grammars with left recursion.
In such cases a simple top-down parser will be
incomplete, and a left corner parser will resort to buffering
the input (so won't be fully word-by-word).
M\&C illustrate the problem by considering
the fragment {\it Mary thinks John}.
This has a small number of possible semantic representations (the
exact number depending upon
the grammar) e.g.
\begin{quote}
{\small{
$\lambda$P.thinks(mary,P(john)) \\
$\lambda$P.$\lambda$Q. Q(thinks(mary,P(john))) \\
$\lambda$P.$\lambda$R. (R($\lambda$x.thinks(x,P(john))))(mary)
}}
\end{quote}
The second representation is appropriate if the sentence finishes with
a sentential modifier. The third allows there
to be a verb phrase modifier.

If the semantic representation is to be read off syntactic structure,
then the parser must provide a single syntax tree (possibly with empty
nodes). However, there are actually any number of such syntax trees
corresponding to, for example, the first semantic representation,
since the {\bf np} and the {\bf s} can be arbitrarily far apart.  The
following tree is suitable for the sentence {\it Mary thinks John shaves}
but not for e.g.\ {\it Mary thinks John coming here was a mistake}.
\begin{verbatim}
             s
            / \
          np  vp
        Mary  / \
             v   s
        thinks  / \
               np  vp^
              John
\end{verbatim}
M\&C suggest various possibilities for packing the partial syntax
trees, including using Tree Adjoining Grammar (Joshi 1987) or
Description Theory (Marcus et al. 1983).  One further possibility
is to choose a single syntax tree, and to use
destructive tree operations later in the parse\footnote{This might
turn out to be similar to one view of Tree Adjoining Grammar, where
adjunction adds into a pre-existing well-formed tree structure. It is
also closer to some methods for incremental adaptation of discourse
structures, where additions are allowed to the right-frontier of a
tree structure (e.g. Polanyi and Scha 1984). There are however
problems with this kind of approach when features are considered (see
e.g.\ Vijay-Shanker 1992).}.

The approach which we will adopt here is based on Milward (1992,
1994).  Partial syntax trees can be regarded as performing two
main roles. The first is to provide syntactic information which guides
how the rest of the sentence can be integrated into the tree. The
second is to provide a basis for a semantic representation. The first
role can be captured using syntactic {\it types}, where each type
corresponds to a potentially infinite number of partial syntax
trees. The second role can be captured by the parser constructing semantic
representations directly. The general processing model therefore consists
of transitions of the form:
\begin{center}
$\begin{array}{c}
{\mbox{Syntactic type}}_{i} \\
{\mbox{Semantic rep}}_{i}
\end{array}$
$\rightarrow$
$\begin{array}{c}
{\mbox{Syntactic type}}_{i+1} \\
{\mbox{Semantic rep}}_{i+1} \\
\end{array}$
\end{center}
This provides a state-transition or {\it dynamic} model of processing,
with each state being a pair of a syntactic type and a semantic value.

The main difference between our approach and that of Milward (1992,
1994) is that it is based on a more expressive grammar formalism,
Applicative Categorial Grammar, as opposed to Lexicalised Dependency
Grammar. Applicative Categorial Grammars allow categories to have
arguments which are themselves functions (e.g. {\bf very} can be
treated as a function of a function, and given the type {\bf
  (n/n)/(n/n)} when used as an adjectival modifier). The ability to
deal with functions of functions has advantages in enabling more
elegant linguistic descriptions, and in providing one kind of robust
parsing: the parser never fails until the last word, since there could
always be a final word which is a function over all the constituents
formed so far.  However, there is a corresponding problem of far
greater non-determinism, with even unambiguous words allowing many
possible transitions. It therefore becomes crucial to either perform
some kind of ambiguity packing, or language {\it tuning}. This will be
discussed in the final section of the paper.

\section{Applicative Categorial Grammar}

Applicative Categorial Grammar is the most basic form of Categorial
Grammar, with just a single combination rule corresponding to function
application. It was first applied to linguistic description by
Adjukiewicz and Bar-Hillel in the 1950s. Although it is still used for
linguistic description (e.g. Bouma and van Noord, 1994), it has been
somewhat overshadowed in recent years by HPSG (Pollard and Sag 1994),
and by Lambek Categorial Grammars (Lambek 1958). It is therefore worth
giving some brief indications of how it fits in with these
developments.

The first directed Applicative CG was proposed by Bar-Hillel
(1953). Functional types included a list of arguments to the left, and
a list of arguments to the right.  Translating Bar-Hillel's notation
into a feature based notation similar to that in HPSG (Pollard and Sag
1994), we obtain the following category for a ditransitive verb such
as {\it put}:
\begin{quote}
\alexcata{\myvalue{\it s}}
         {\langle{\myvalue{\it np}}\rangle}
         {\langle{\myvalue{\it np, pp}}\rangle}
\end{quote}
The list of arguments to the left are gathered under the feature, {\bf l},
and those to the right, an {\bf np} and a {\bf pp} in that order,
under the feature {\bf r}.

Bar-Hillel employed a single application rule, which corresponds to
the following: \\[2mm]
\hspace*{2mm} {\it L$_{n}$ \ldots L$_{1}$}
\alexcata{\myvalue{\it X}}
         {\langle{\myvalue{\it L$_{1}$ \ldots L$_{n}$}}\rangle}
         {\langle{\myvalue{\it R$_{1}$ \ldots R$_{n}$}}\rangle}
{\it R$_{1}$ \ldots R$_{n}$}
$\Rightarrow$ X
\\[2mm]
The result was a system which comes very close to the formalised
dependency grammars of Gaifman (1965) and Hays (1964). The only real
difference is that Bar-Hillel allowed arguments to themselves be
functions.  For example, an adverb such as {\it slowly} could be given
the type\footnote{The reformulation is not entirely faithful here to
Bar-Hillel, who used a slightly problematic `double slash' notation
for functions of functions.}
\begin{quote}
\alexcata{\myvalue{\it s}}
         {\langle{\myvalue{\it np}}\rangle}
         {\langle\alexcata{\myvalue{\it s}}
                          {\langle{\myvalue{\it np}}\rangle}
                          {\langle\rangle}\rangle}
\end{quote}
An unfortunate aspect of Bar-Hillel's first system was that the
application rule only ever resulted in a primitive type. Hence,
arguments with functional types had to correspond to single lexical
items: there was no way to form the type {\bf
  np$\backslash$s}\footnote{Lambek notation (Lambek 1958).} for a
non-lexical verb phrase such as {\it likes Mary}.

Rather than adapting the Application Rule to allow functions to be
applied to one argument at a time, Bar-Hillel's second system (often
called AB Categorial Grammar, or Adjukiewicz/Bar-Hillel CG, Bar-Hillel
1964) adopted a `Curried' notation, and this has been adopted by most
CGs since.  To represent a function which requires an {\bf np} on the
left, and an {\bf np} and a {\bf pp} to the right, there is a choice
of the following three types using Curried notation:
\begin{quote}
np$\backslash$((s/pp)/np) \\
(np$\backslash$(s/pp))/np \\
((np$\backslash$s)/pp)/np
\end{quote}
Most CGs either choose the third of these (to give a {\bf vp} structure), or
include a rule of Associativity which means that the types are
interchangeable (in the Lambek Calculus, Associativity is a
consequence of the calculus, rather than being specified
separately).

The main impetus to change Applicative CG came from the work of Ades and
Steedman (1982). Ades and Steedman noted that the use of function composition
allows CGs to deal with unbounded dependency constructions. Function
composition enables a function to be applied to its argument, even if that
argument is incomplete e.g.
\begin{quote}
s/pp + pp/np $\rightarrow$ s/np
\end{quote}
This allows peripheral extraction, where the `gap' is at the start or
the end of e.g. a relative clause. Variants of the composition rule
were proposed in order to deal with non-peripheral extraction, but
this led to unwanted effects elsewhere in the grammar (Bouma 1987).
Subsequent treatments of non-peripheral extraction based on the
Lambek Calculus (where standard composition is built in: it is a rule
which can be proven from the calculus) have either introduced an
alternative to the forward and backward slashes i.e.\ / and
$\backslash$ for normal args, $\uparrow$ for wh-args (Moortgat 1988),
or have introduced so called modal operators on the
wh-argument (Morrill et al.\ 1990).  Both techniques can be thought of
as marking the wh-arguments as requiring special treatment, and
therefore do not lead to unwanted effects elsewhere in the grammar.

However, there are problems with having just composition, the most
basic of the non-applicative operations.  In CGs which contain
functions of functions (such as {\it very}, or {\it slowly}), the
addition of composition adds both new analyses of sentences, and new
strings to the language. This is due to the fact that composition can
be used to form a function, which can then be used as an argument to a
function of a function.  For example, if the two types, {\bf n/n} and
{\bf n/n} are composed to give the type {\bf n/n}, then this can be
modified by an adjectival modifier of type {\bf (n/n)/(n/n)}.  Thus,
the noun {\it very old dilapidated car} can get the unacceptable
bracketing, [[very [old dilapidated]] car]. Associative CGs with
Composition, or the Lambek Calculus also allow strings such as {\it
  boy with the} to be given the type {\bf n/n} predicting {\it very
  boy with the car} to be an acceptable noun.  Although individual
examples might be possible to rule out using appropriate features, it
is difficult to see how to do this in general whilst retaining a
calculus suitable for incremental interpretation.

If wh-arguments need to be treated specially anyway (to deal with
non-peripheral extraction), and if composition as a general rule is
problematic, this suggests we should perhaps return to grammars which
use just Application as a general operation, but have a special
treatment for wh-arguments.  Using the non-Curried notation of
Bar-Hillel, it is more natural to use a separate wh-list than to
mark wh-arguments individually. For example, the category
appropriate for relative clauses with a noun phrase gap would be:
\begin{quote}
\alexcatw{\myvalue{\it s}}
         {\langle\rangle}
         {\langle\rangle}
         {\langle{\myvalue{\it np}}\rangle}
\end{quote}
It is then possible to specify operations which act as purely
applicative operations with respect to the left and right arguments
lists, but more like composition with respect to the wh-list.  This is
very similar to the way in which wh-movement is dealt with in GPSG
(Gazdar et al.\ 1985) and HPSG, where wh-arguments are treated using
slash mechanisms or feature inheritance principles which correspond
closely to function composition.

Given that our arguments have produced a categorial grammar which
looks very similar to HPSG, why not use HPSG rather than Applicative CG?
The main reason is that Applicative CG is a much simpler formalism,
which can be given a very simple syntax semantics interface,
with function application in syntax mapping to function application in
semantics\footnote{One area where application based approaches to
  semantic combination gain in simplicity over unification based
  approaches is in providing semantics for functions of functions.
  Moore (1989) provides a treatment of functions of functions in a
  unification based approach, but only by explicitly incorporating
  lambda expressions. Pollard and Sag (1994) deal with some functions
  of functions, such as non-intersective adjectives, by explicit set
  construction.}$^{,}$\footnote{As discussed above, wh-movement requires
  something more like composition than application.  A simple syntax
  semantics interface can be retained if the same operation is used in
  both syntax and semantics.  Wh-arguments can be treated as similar
  to other arguments i.e.\ as lambda abstracted in the semantics.  For
  example, the fragment: {\it John found a woman who Mary} can be
  given the semantics {\bf $\lambda$P.$\exists$x. woman(x) \&
    found(john,x) \& P(mary,x)}, where {\bf P} is a function from a
  left argument {\it Mary} of type {\bf e} and a wh-argument, also of
  type {\bf e}.}.  This in turn makes it relatively easy to provide
proofs of soundness and completeness for an incremental parsing
algorithm.  Ultimately, some of the techniques developed here should be
able to be extended to more complex formalisms such as HPSG.

\section{AB Categorial grammar with Associativity (AACG)}

In this section we define a grammar similar to Bar-Hillel's first
grammar. However, unlike Bar-Hillel, we allow one argument to be
absorbed at a time.  The resulting grammar is equivalent to AB
Categorial Grammar plus associativity.

The categories of the grammar are defined as follows:
\begin{enumerate}
\item If {\bf X} is a syntactic type (e.g.\ s, np), then
\alexcata{\myvalue{\it X}}
               {\langle\rangle}
               {\langle\rangle}
is a category.
\item
If {\bf X} is a syntactic type, and {\bf L} and {\bf R} are
lists of categories, then \\[1mm]
\alexcata{\myvalue{\it X}}
              {\myvalue{\it L}}
              {\myvalue{\it R}} is a category.
\end{enumerate}
Application to the right is defined by the rule\footnote{`$\bullet$'
  is list concatenation e.g.
  $\langle$np$\rangle\bullet\langle$s$\rangle$ equals
  $\langle$np,s$\rangle$.}:
\begin{quote}
\alexcata{\myvalue{\it X}}
              {\myvalue{\it L}}
              {\myvalue{\it $\langle$R$_{1}\rangle\bullet$R}}
$+$
R$_{1}$  $\Rightarrow$ \alexcata{\myvalue{\it X}}
              {\myvalue{\it L}}
              {\myvalue{\it R}}
\end{quote}
Application to the left is defined by the rule:
\begin{quote}
L$_{1}$ $+$
\alexcata{\myvalue{\it X}}
              {\myvalue{\it $\langle$L$_{1}\rangle\bullet$L}}
              {\myvalue{\it R}}
$\Rightarrow$
\alexcata{\myvalue{\it X}}
              {\myvalue{\it L}}
              {\myvalue{\it R}}
\end{quote}
The basic grammar provides some spurious derivations, since sentences
such as {\it John likes Mary} can be bracketed as either {\it ((John
  likes) Mary)} or {\it (John (likes Mary))}. However, we will see
that these spurious derivations do not translate into spurious
ambiguity in the parser, which maps from strings of words directly to
semantic representations.

\begin{figure*}[t]
State-Application: \\[1mm]
$\begin{array}{l} \alexcath{\myvalue{\it Y}} {\langle\rangle}
{\langle\alexcath{\myvalue{\it X}}
                 {\myvalue{\it L$_{0}$}}
                 {\myvalue{\it R$_{0}$}}
                 {\myvalue{\it H$_{0}$}}
\rangle\bullet{\myvalue{\it R$_{2}$}}}
{\langle\rangle} \\
{\mbox{\small{\bf F}}}
\end{array}$
$\stackrel{\mbox{\small ``W''}}{\rightarrow}$
$\begin{array}{l}
\alexcath{\myvalue{\it Y}}
         {\langle\rangle}
         {\myvalue{\it R$_{1}\bullet$R$_{2}$}}
         {\langle\rangle}
\hspace{55mm}
{\mbox{{\small{where}} \hspace{1mm} W:}}
\begin{array}{l}
\alexcath{\myvalue{\it X}} {\myvalue{\it
L$_{0}$}} {\myvalue{\it R$_{1}\bullet$R$_{0}$}} {\langle\rangle} \\
{\mbox{\small{\bf G}}}
\end{array} \\
{\mbox{\small{\bf $\lambda${\bf r}$_{1}$. F(G({\bf r}$_{1}$))}}}
\end{array}$
\\[4mm]
State-Prediction: \\[-5mm]
$\begin{array}{l}
\alexcath{\myvalue{\it Y}}
         {\langle\rangle}
         {\langle\alexcath{\myvalue{\it X}}
                       {\myvalue{\it L$_{1}\bullet$L$_{0}$}}
                       {\myvalue{\it R$_{0}$}}
                       {\myvalue{\it L$_{1}\bullet$H$_{0}$}}
\rangle\bullet{\myvalue{\it R$_{2}$}}}
{\langle\rangle}
\\
{\mbox{\small {\bf F}}}
\end{array}$
$\stackrel{\mbox{\small ``W''}}{\rightarrow}$
$\begin{array}{l}
\alexcath{\myvalue{\it Y}}
         {\langle\rangle}
{\myvalue{\it R$_{1}$}\bullet\langle{\alexcath{\myvalue{\it X}}
                                 {\langle{\alexcath{\myvalue{\it Z}}
                                                   {\myvalue{\it L}}
                                                   {\myvalue{\it R}}
                                                   {\langle\rangle}}
\rangle\bullet{\myvalue{\it L$_{0}$}}}
{\myvalue{\it R$_{0}$}}
{\langle{\alexcath{\myvalue{\it Z}}
                                                   {\myvalue{\it L}}
                                                   {\myvalue{\it R}}
                                                   {\langle\rangle}}
\rangle\bullet{\myvalue{\it H$_{0}$}}}
}\rangle\bullet{\myvalue{\it R$_{2}$}}}
{\langle\rangle} \hspace*{15mm}
{\mbox{{\small where} \hspace{1mm} W:}}
\begin{array}{l}
\alexcath{\myvalue{\it Z}}
                                {\myvalue{\it L$_{1}\bullet$L}}
                                {\myvalue{\it R$_{1}\bullet$R}}
                                {\langle\rangle}\\
{\mbox{\small {\bf G}}}
\end{array}
\\
{\mbox{\small{\bf $\lambda${\bf r}$_{1}$.($\lambda$h.
F($\lambda${\bf l}$_{1}$. (h( $\lambda${\bf r}
(((G {\bf r}$_{1}$){\bf r}){\bf l}$_{1}$)))))}}}
\end{array}$
\caption{Transition Rules}
\end{figure*}

\section{An Incremental Parser}

Most parsers which work left to right along an input string can be
described in terms of state transitions i.e.\ by rules which say how
the current parsing state (e.g.\ a stack of categories, or a chart)
can be transformed by the next word into a new state.  Here this will
be made particularly explicit, with the parser described in terms of
just two rules which take a state, a new word and create a new
state\footnote{This approach is described in greater detail
in Milward (1994), where
parsers are specified formally in terms of their dynamics.}. There are
two unusual features. Firstly, there is nothing equivalent to a stack
mechanism: at all times the state is characterised by a single
syntactic type, and a single semantic value, not by some stack of
semantic values or syntax trees which are waiting to be connected
together. Secondly, all transitions between states occur on the input
of a new word: there are no `empty' transitions (such as the reduce
step of a shift-reduce parser).

The two rules, which are given in Figure 1\footnote{{\bf L$_{i}$}, {\bf
    R$_{i}$}, {\bf H$_{i}$} are lists of categories. {\bf l$_{i}$} and
  {\bf r$_{i}$} are lists of variables, of the same length as the
  corresponding {\bf L$_{i}$} and {\bf R$_{i}$}.},
are difficult to
understand in their most general form. Here we will work upto the
rules gradually, by considering which kinds of rules we might need in
particular instances. Consider the following pairing of sentence
fragments with their simplest possible CG type:
\begin{quote}
Mary thinks: s/s \\
Mary thinks John: s/(np$\backslash$s) \\
Mary thinks John likes: s/np \\
Mary thinks John likes Sue: s
\end{quote}
Now consider taking each type as a description of the state that the
parser is in after absorbing the fragment. We obtain a sequence
of transitions as follows:
\begin{quote}
s/s $\stackrel
{\mbox{\small ``John''}}{\rightarrow}$
s/(np$\backslash$s)
$\stackrel{\mbox{\small ``likes''}}{\rightarrow}$
s/np
$\stackrel{\mbox{\small ``Sue''}}{\rightarrow}$
s
\end{quote}
If an embedded sentence such as {\it John likes Sue} is a mapping from an
{\bf s/s} to an {\bf s}, this suggests that it might be possible to
treat all sentences as mapping from some category expecting an {\bf s}
to that category i.e.\ from {\bf X/s} to {\bf X}. Similarly, all noun
phrases might be treated as mappings from an {\bf X/np} to an {\bf X}.

Now consider individual transitions.  The simplest of these is
where the type of argument expected by the state is matched by the
next word i.e.
\begin{quote}
s/np $\stackrel{\mbox{\small ``Sue''}}{\rightarrow}$ s \hspace*{1cm}
{\small where:} \hspace*{2mm} Sue: np
\end{quote}
This can be generalised to the following rule, which is similar
to Function Application in standard CG\footnote{It differs in not being a rule
of grammar: here the functor is a state category and the argument is a
lexical category. In standard CG function application, the functor and
argument can correspond to a word or a phrase.}
\begin{quote}
X/Y  $\stackrel{\mbox{\small ``W''}}{\rightarrow}$ X \hspace*{1cm}
{\small where:} \hspace*{2mm} W: Y
\end{quote}
A similar transition occurs for {\it likes}. Here an {\bf np$\backslash$s} was
expected, but {\it likes} only provides part of this: it requires an
{\bf np} to the right to form an {\bf np$\backslash$s}. Thus after {\it
  likes}
is absorbed the state category will need to expect an {\bf np}.
The rule required is similar to Function Composition in CG i.e.
\begin{quote}
X/Y  $\stackrel{\mbox{\small ``W''}}{\rightarrow}$ X/Z \hspace*{1cm}
{\small where:} \hspace*{2mm} W: Y/Z
\end{quote}

\begin{figure*}[t]
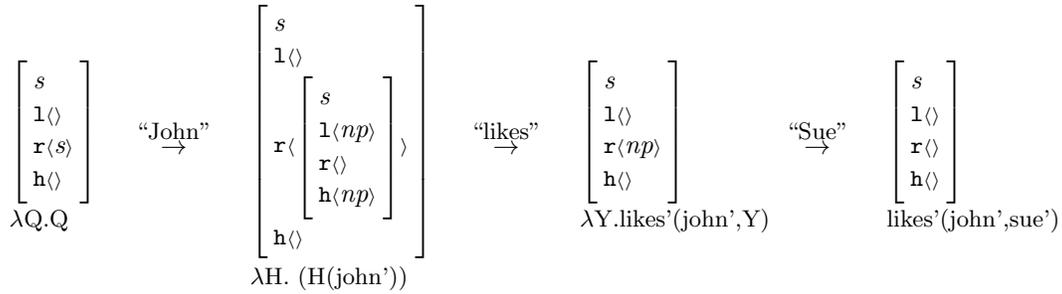

\begin{center}
$\begin{array}{l}
\alexcath{\myvalue{\it s}}
         {\langle\rangle}
         {\langle{\myvalue{\it s}}\rangle}
         {\langle\rangle} \\
{\footnotesize{\mbox{$\lambda$Q.Q}}}
\end{array}$
\hspace*{1mm} $\stackrel{\mbox{\small ``John''}}{\rightarrow}$
\hspace*{1mm}
$\begin{array}{l}
\alexcath{\myvalue{\it s}}
                   {\langle\rangle}
                   {\langle{\alexcath{\myvalue{\it s}}
                                     {\langle{\myvalue{\it np}}\rangle}
                                     {\langle\rangle}
                                     {\langle{\myvalue{\it np}}\rangle}
                                      }\rangle}
                   {\langle\rangle} \\
{\footnotesize{\mbox{$\lambda$H. (H(john'))}}}
\end{array}$
\hspace*{1mm} $\stackrel{\mbox{\small ``likes''}}{\rightarrow}$
\hspace*{1mm}
$\begin{array}{l}
\alexcath{\myvalue{\it s}}
         {\langle\rangle}
         {\langle{\myvalue{\it np}}\rangle}
         {\langle\rangle} \\
{\footnotesize{\mbox{$\lambda$Y.likes'(john',Y)}}}
\end{array}$
\hspace*{-2mm} $\stackrel{\mbox{\small ``Sue''}}{\rightarrow}$
\hspace*{1mm}
$\begin{array}{l}
\alexcath{\myvalue{\it s}}
         {\langle\rangle}
         {\langle\rangle}
         {\langle\rangle} \\
{\footnotesize{\mbox{likes'(john',sue')}}}
\end{array}$
\end{center}
\caption{Possible state transitions}
\end{figure*}

\noindent
Considering this informally in terms of tree structures, what is
happening is the replacement of an empty node in a partial tree by a
second partial tree i.e.
\begin{verbatim}
    X                           X
   / \                         / \
  U   Y^    +     Y     =>    U   Y
                 / \             / \
                V   Z^          V   Z^
\end{verbatim}
The two rules specified so far need to be further generalised to allow
for the case where a lexical item has more than one argument (e.g. if
we replace {\it likes} by a di-transitive such as {\it gives} or a
tri-transitive such as {\it bets}).  This is relatively trivial using
a non-curried notation similar to that used for AACG. What we obtain
is the single rule of
State-Application, which corresponds to application when
the list of arguments, {\bf R$_{1}$}, is empty, to function
composition when {\bf R$_{1}$} is of length one, and to n-ary
composition when {\bf R$_{1}$} is of length {\bf n}.  The only change
needed from AACG notation is the inclusion of an extra feature
list, the {\bf h} list, which stores information about which arguments
are waiting for a head (the reasons for this will be explained later).
The lexicon is identical to that for a standard AACG, except for
having h-lists which are always set to empty.

Now consider the first transition. Here a sentence was expected, but
what was encountered was a noun phrase, {\it John}.
The appropriate rule in CG notation would be:
\begin{quote}
X/Y $\stackrel{\mbox{\small ``W''}}{\rightarrow}$
X/(Z$\backslash$Y) \hspace*{4mm}
{\small where:} \hspace*{2mm} W: Z
\end{quote}
This rule states that if looking for a {\bf Y} and get a {\bf Z} then
look for a {\bf Y} which is missing a {\bf Z}. In tree structure terms
we have:
\begin{verbatim}

    X                         X
   / \                       / \
  U   Y^   +    Z     =>    U   Y
                               / \
                              Z  Z\Y^
\end{verbatim}
The rule of State-Prediction is obtained by further generalising to
allow the lexical item to have missing arguments, and for the expected
argument to have missing arguments.

State-Application and State-Prediction together provide the basis of a
sound and complete parser\footnote{The parser accepts the same strings
  as the grammar and assigns them the same semantic values. This is
  slightly different from the standard notion of soundness and
  completeness of a parser, where the parser accepts the same strings
  as the grammar and assigns them the same syntax trees.}. Parsing of
sentences is achieved by starting in a state expecting a sentence, and
applying the rules non-deterministically as each word is input. A
successful parse is achieved if the final state expects no more
arguments. As an example, reconsider the string {\it John likes Sue}.
The sequence of transitions corresponding to {\it
  John likes Sue} being a sentence, is given in Figure 2.
The transition on encountering
{\it John} is deterministic: State-Application cannot apply, and
State-Prediction can only be instantiated one way. The result is
a new state expecting an argument which, given an {\bf np} could
give an {\bf s} i.e.\ an {\bf np$\backslash$s}.

\begin{figure*}[t]
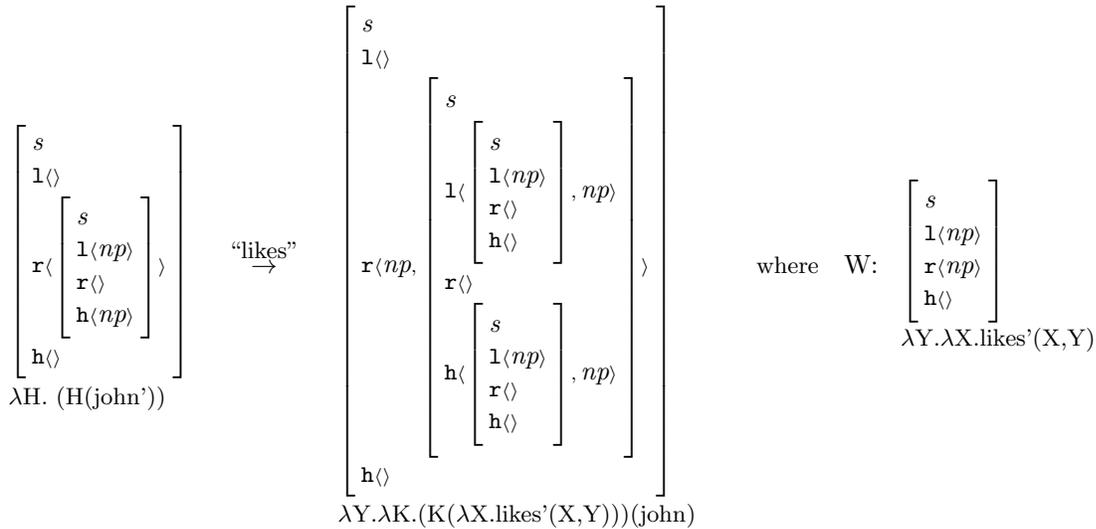

$\begin{array}{l}
\alexcath{\myvalue{\it s}}
                   {\langle\rangle}
                   {\langle{\alexcath{\myvalue{\it s}}
                                     {\langle{\myvalue{\it np}}\rangle}
                                     {\langle\rangle}
                                     {\langle{\myvalue{\it np}}\rangle}
                                      }\rangle}
                   {\langle\rangle} \\
{\small{\mbox{$\lambda$H. (H(john'))}}}
\end{array}$
\hspace*{1mm} $\stackrel{\mbox{\small ``likes''}}{\rightarrow}$
\hspace*{1mm}
$\begin{array}{l}
\alexcath{\myvalue{\it s}}
                   {\langle\rangle}
                   {\langle{\myvalue{\it np}},\alexcath{\myvalue{\it s}}
                                     {\langle{
\alexcath{\myvalue{\it s}}
         {\langle{\myvalue{\it np}}\rangle}
         {\langle\rangle}
         {\langle\rangle},
\myvalue{\it np}}
\rangle}
                 {\langle\rangle}
                 {\langle{\alexcath{\myvalue{\it s}}
                                   {\langle{\myvalue{\it np}}\rangle}
                                   {\langle\rangle}
                                   {\langle\rangle}, \myvalue{\it np}}
\rangle}
                                      \rangle}
                   {\langle\rangle} \\
{\small{\mbox{$\lambda$Y.$\lambda$K.(K($\lambda$X.likes'(X,Y)))(john)}}}
\end{array}$
\hspace*{4mm}
{\small where} \hspace{1mm} W:
$\begin{array}{l}
\alexcath{\myvalue{\it s}}
                                {\langle{\myvalue{\it np}}\rangle}
                                {\langle{\myvalue{\it np}}\rangle}
                                {\langle\rangle}\\
{\mbox{\small $\lambda$Y.$\lambda$X.likes'(X,Y)}}
\end{array}$
\caption{Example instantiation of State-Prediction}
\end{figure*}

The transition on input of {\it likes} is non-deterministic.
State-Application can apply, as in Figure 2.  However,
State-Prediction can also apply, and can be instantiated in four ways
(these correspond to different ways of cutting up the left and right
subcategorisation lists of the lexical entry, {\it likes}, i.e.\ as
$\langle$np$\rangle\bullet\langle\rangle$ or
$\langle\rangle\bullet\langle$np$\rangle$).  One possibility
corresponds to the prediction of an {\bf s$\backslash$s} modifier, a
second to the prediction of an {\bf
  (np$\backslash$s)$\backslash$(np$\backslash$s)} modifier (i.e.\  a
verb phrase modifier), a third to there being a function which takes
the subject and the verb as separate arguments, and the fourth
corresponds to there being a function which requires an {\bf s/np}
argument. The second
of these is perhaps the most interesting, and is given in Figure 3.
It is the choice of this particular transition at this point which
allows verb phrase modification, and hence, assuming the next word is
{\it Sue}, an implicit bracketing of the string fragment as {\it (John
(likes Sue))}. Note that if State-Application is chosen, or the
first of the State-Prediction possibilities, the fragment {\it John
likes Sue} retains a flat structure.  If there is to be no
modification of the verb phrase, no verb phrase structure is
introduced. This relates to there being no spurious ambiguity: each
choice of transition has {\it semantic} consequences; each choice
affects whether a particular part of the semantics is to be modified
or not.

Finally, it is worth noting why it is necessary to use h-lists. These
are needed to distinguish between cases of real functional arguments
(of functions of functions), and functions formed by
State-Prediction.  Consider the following trees, where the {\bf
  np$\backslash$s} node is empty.
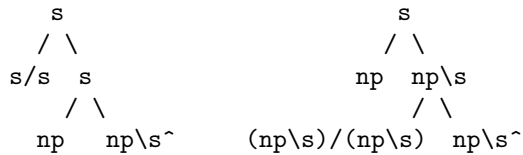
\begin{verbatim}
    s                        s
   / \                      / \
 s/s  s                   np  np\s
     / \                      / \
   np   np\s^     (np\s)/(np\s)  np\s^
\end{verbatim}
Both trees have the same syntactic type, however in the first case we
want to allow for there to be an {\bf s$\backslash$s} modifier of the
lower {\bf s}, but not in the second. The headed list distinguishes
between the two cases, with only the first having an {\bf np} on its
headed list, allowing prediction of an {\bf s} modifier.

\section{Parsing Lexicalised Grammars}

When we consider full sentence processing, as opposed to incremental
processing, the use of lexicalised grammars has a major advantage over
the use of more standard rule based grammars.  In processing a
sentence using a lexicalised formalism we do not have to look at the
grammar as a whole, but only at the grammatical information indexed by
each of the words. Thus increases in the size of a grammar don't
necessarily effect efficiency of processing, provided the increase in
size is due to the addition of new words, rather than increased
lexical ambiguity.  Once the full set of possible lexical entries for
a sentence is collected, they can, if required, then be converted back
into a set of phrase structure rules (which should correspond to a
small subset of the rule based formalism equivalent to the whole
lexicalised grammar), before being parsing with a standard algorithm
such as Earley's (Earley 1970).

In incremental parsing we cannot predict which words will appear in
the sentence, so cannot use the same technique.  However, if
we are to base a parser on the rules given above, it would seem that we
gain further. Instead of grammatical information being localised to
the sentence as a whole, it is localised to a particular word in
its particular context: there is no need to consider a {\bf pp} as a start
of a sentence if it occurs at the end, even if there is a verb with an
entry which allows for a subject {\bf pp}.

However there is a major problem. As we noted in the last paragraph,
it is the nature of parsing incrementally that we don't know what
words are to come next. But here the parser doesn't even use the
information that the words are to come from a lexicon for a particular
language.  For example, given an input of 3 nps, the parser will
happily create a state expecting 3 nps to the left. This might be a
likely state for say a head final language, but an unlikely state for
a language such as English. Note that incremental interpretation will
be of no use here, since the semantic representation should be no more
or less plausible in the different languages. In practical terms, a
naive interactive parallel Prolog implementation on a current
workstation fails to be interactive in a real sense after about 8
words\footnote{This result should however be treated with some caution: in
this implementation there was no attempt to perform any packing of
different possible transitions, and the algorithm has exponential
complexity. In contrast, a packed recogniser based on a similar, but
much simpler, incremental
parser for Lexicalised Dependency Grammar has {\it
  O}(n$^{3}$) time complexity (Milward 1994) and good practical
performance, taking a couple of seconds on 30 word sentences.}.

What seems to be needed is some kind of {\it language
tuning}\footnote{The usage of the term {\it language tuning} is perhaps
  broader here than its use in the psycholinguistic literature to
  refer to different structural preferences between languages e.g. for
  high versus low attachment (Mitchell et al.\ 1992).}.  This could be
in the nature of fixed restrictions to the rules e.g.\ for English we
might rule out uses of prediction when a noun phrase is encountered,
and two already exist on the left list.
A more appealing alternative is to base the tuning on statistical methods.
This could be achieved by running the parser over
corpora to provide probabilities of particular transitions given
particular words. These transitions would capture the likelihood of a
word having a particular part of speech, and the probability of a
particular transition being performed with that part of speech.

There has already been some early work done on providing statistically
based parsing using transitions between recursively structured
syntactic categories (Tugwell 1995)\footnote{Tugwell's approach does
however differ in that the state transitions are not limited by the
rules of State-Prediction and State-Application. This has advantages
in allowing the grammar to learn phenomena such as heavy NP shift, but
has the disadvantage of suffering from greater sparse data problems.
A compromise system using the rules here, but allowing reordering of
the r-lists might be preferable.}.  Unlike a simple Markov
process, there are a potentially infinite number of states, so there
is inevitably a problem of sparse data. It is therefore necessary to
make various generalisations over the states, for example by ignoring
the {\bf R$_{2}$} lists.

The full processing model can then be either serial, exploring the most
highly ranked transitions first (but allowing backtracking
if the semantic plausibility of the current interpretation drops too
low), or ranked parallel, exploring just the {\bf n} paths ranked
highest according to the transition probabilities and semantic
plausibility.

\section{Conclusion}

The paper has presented a method for providing interpretations word by
word for basic Categorial Grammar. The final section contrasted
parsing with lexicalised and rule based grammars, and argued that
statistical language tuning is particularly suitable for incremental,
lexicalised parsing strategies.

\section*{References}

\noindent
Ades, A.\ \& Steedman, M.: 1972, `On the Order of Words', {\it Linguistics
\& Philosophy} 4, 517-558.\\[2mm]
Bar-Hillel, Y.: 1953, `A Quasi-Arithmetical Notation for Syntactic
Description', {\it Language} 29, 47-58.\\[2mm]
Bar-Hillel, Y.: 1964, {\it Language \& Information: Selected Essays
on Their Theory \& Application}, Addison-Wesley.\\[2mm]
Bouma, G.: 1987, `A Unification-Based Analysis of Unbounded Dependencies',
in {\it Proceedings of the 6th Amsterdam Colloquium}, ITLI, University
of Amsterdam. \\[2mm]
Bouma, G. \& van Noord, G.: 1994, `Constraint-Based Categorial Grammar',
in {\it Proceedings of the 32nd ACL}, Las Cruces, U.S.A. \\[2mm]
Earley, J.: 1970, `An Efficient Context-free Parsing Algorithm', {\it
ACM Communications} 13(2), 94-102.\\[2mm]
Gaifman, H.: 1965, `Dependency Systems \& Phrase Structure Systems',
{\it Information \& Control} 8: 304-337.\\[2mm]
Gazdar, G., Klein, E., Pullum, G.K., \& Sag, I.A.: 1985, {\it
Generalized Phrase Structure Grammar}, Blackwell, Oxford.
\\[2mm]
Hays, D.G.: 1964, `Dependency Theory: A Formalism \& Some Observations',
{\it Language} 40, 511-525. \\[2mm]
Joshi, A.K.: 1987, `An Introduction to Tree Adjoining Grammars', in
Manaster-Ramer (ed.), {\it Mathematics of Language}, John Benjamins,
Amsterdam.
\\[2mm]
Lambek, J.: 1958, `The Mathematics of Sentence Structure', {\it American
Mathematical Monthly} 65, 154-169.
\\[2mm]
Marcus, M., Hindle, D., \& Fleck, M.: 1983, `D-Theory: Talking about
Talking about Trees', in {\it Proceedings of the 21st ACL}, Cambridge, Mass.
\\[2mm]
Marslen-Wilson, W.: 1973, `Linguistic Structure \& Speech Shadowing
at Very Short Latencies', {\it Nature} 244, 522-523.
\\[2mm]
Milward, D.: 1992, `Dynamics, Dependency Grammar \& Incremental
Interpretation',
in {\it Proceedings of COLING 92}, Nantes, vol 4, 1095-1099.\\[2mm]
Milward, D. \& Cooper, R.: 1994, `Incremental Interpretation: Applications,
Theory \& Relationship to Dynamic Semantics', in {\it Proceedings of
COLING 94}, Kyoto, Japan, 748-754. \\[2mm]
Milward, D.: 1994, `Dynamic Dependency Grammar', to appear in {\it
Linguistics \& Philosophy} 17, 561-605.\\[2mm]
Mitchell, D.C., Cuetos, F., \& Corley, M.M.B.: 1992, `Statistical versus
linguistic determinants of parsing bias: cross-linguistic evidence'.
Paper presented at the 5th Annual CUNY Conference on Human Sentence
Processing, New York. \\[2mm]
Moore, R.C.: 1989, `Unification-Based Semantic Interpretation', in
{\it Proceedings of the 27th ACL}, Vancouver. \\[2mm]
Moortgat, M.: 1988, {\it Categorial Investigations: Logical \& Linguistic
Aspects of the Lambek Calculus}, Foris, Dordrecht. \\[2mm]
Morrill, G., Leslie, N., Hepple, M. \& Barry,G.: 1990, `Categorial
Deductions \& Structural Operations', in Barry, G. \&
Morrill, G. (eds.), {\it Studies in Categorial
Grammar}, Edinburgh Working Papers in Cognitive Science, 5. \\[2mm]
Polanyi, L. \& Scha, R.: 1984, `A Syntactic Approach to Discourse Semantics',
in {\it Proceedings of COLING 84}, Stanford, 413-419. \\[2mm]
Pollard, C. \& Sag, I.A.: 1994, {\it Head-Driven Phrase Structure Grammar},
University of Chicago Press \& CSLI Publications, Chicago.\\[2mm]
Pulman, S.G.: 1986, `Grammars, Parsers, \& Memory Limitations', {\it
Language
\& Cognitive Processes} 1(3), 197-225.\\[2mm]
Spivey-Knowlton, M., Sedivy, J., Eberhard, K., \& Tanenhaus, M.:
1994, `Psycholinguistic Study of the Interaction Between Language \&
Vision', in {\it Proceedings of the 12th National Conference on AI, AAAI-94}.
\\[2mm]
Stabler, E.P.: 1991, `Avoid the Pedestrian's Paradox', in Berwick,
R.C. et al.\ (eds.), {\it Principle-Based
Parsing: Computation \& Psycholinguistics}, Kluwer, Netherlands,
199-237.\\[2mm]
Steedman, M.J.: 1991, `Type-Raising \& Directionality in Combinatory Grammar',
in Proceedings of the 29th ACL, Berkeley, U.S.A. \\[2mm]
Tanenhaus, M.K., Garnsey, S., \& Boland, J.: 1990, `Combinatory
Lexical
Information \& Language Comprehension', in Altmann, G.T.M. {\it
  Cognitive
Models of Speech Processing}, MIT Press, Cambridge Ma.
\\[2mm]
Tugwell, D.: 1995, `A State-Transition Grammar for Data-Oriented
Parsing', in Proceedings of the 7th Conference of the European
  ACL, EACL-95, Dublin, this volume. \\[2mm]
Vijay-Shanker, K.: 1992, `Using Descriptions of Trees in a Tree Adjoining
Grammar', {\it Computational Linguistics} 18(4), 481-517.

\end{document}